\documentclass[sigconf]{acmart}
\usepackage{xcolor}
\usepackage{multirow}
\usepackage{algorithm}
\usepackage{algpseudocode}
\usepackage{graphicx}
\usepackage{subcaption}

\definecolor{mycolor}{RGB}{219,219,219} 

\setcopyright{acmlicensed}
\copyrightyear{2025}
\acmYear{2025}
\acmDOI{XX.XXX}
\acmConference[Conference CIKM '25]{The 34th ACM International Conference on Information and Knowledge Management}{November 10--14,
  2025}{SEOUL, KOREA}
\acmISBN{978-1-4503-XXXX-X/2025/06}

\acmSubmissionID{5760}

\begin{document}

\title{AME: Aligned Manifold Entropy for Robust Vision-Language Distillation}

\author{Guiming Cao \and Yuming Ou }

\renewcommand{\shortauthors}{Cao et al.}


\begin{abstract}
Knowledge distillation is a long-established technique for knowledge transfer, and has regained attention in the context of the recent emergence of large vision-language models (VLMs).
However, vision-language knowledge distillation often requires sufficient training data to achieve robust generalization on samples with ambiguous or boundary-adjacent representations, which are associated with high predictive uncertainty. 
Critically, collecting such large-scale, task-specific data for training is often impractical in real-world scenarios.
To address this major challenge arising from the entanglement of uncertainty and cross-modal feature representation, we propose \textbf{A}ligned \textbf{M}anifold \textbf{E}ntropy for Robust Vision-Language Distillation (AME), aiming to achieve robust generalization under real-world conditions.
AME applies entropy minimization over a reconfigured shared manifold, where multi-modal data (i.e., image and text) are bridged through a pair of projection functions, conducive to structural compression for cross-modal feature representations.
This enables robust knowledge distillation under low-data regimes, while requiring no architectural modifications to the backbone. As a result, it can serve as a plug-and-play module compatible with a wide range of vision-language distillation frameworks.
Notably, our theoretical analysis reveals that integrating knowledge distillation with entropy minimization over the shared manifold leads to a tighter generalization error bound.
Extensive experiments across diverse distillation architectures and training settings demonstrate that AME consistently facilitates robust knowledge distillation, resulting in superior generalization performance across a wide spectrum of downstream tasks.
These findings highlight AME as a general and robust mechanism for knowledge distillation, paving the way for the adaptation of vision-language models to a broad spectrum of real-world scenarios.

\end{abstract}



\received{18 May 2025}
\received[revised]{18 May 2025}
\received[accepted]{18 May 2025}

\maketitle

\section{Introduction}
Vision Language Model (VLM) is a representation learning paradigm that interprets and aligns cross-modal features, signifying a major leap forward over the traditional paradigm in a wide range of multi-modal downstream tasks. 
Among such an emerging paradigm, CLIP~\cite{CLIP} and ALGIN~\cite{ALIGN} stand out as prominent approaches trained on large-scale datasets with 400 million text-image pairs, demonstrating superb performance in zero-shot and few-shot settings.
In this context, knowledge distillation (KD) has gained renewed interest as an effective means of adapting VLMs to a wide spectrum of downstream tasks in the era of multi-modal learning.
\begin{figure}[!t]
  \centering
  
  \begin{minipage}[t]{0.98\linewidth}
    \centering
    \includegraphics[width=\linewidth]{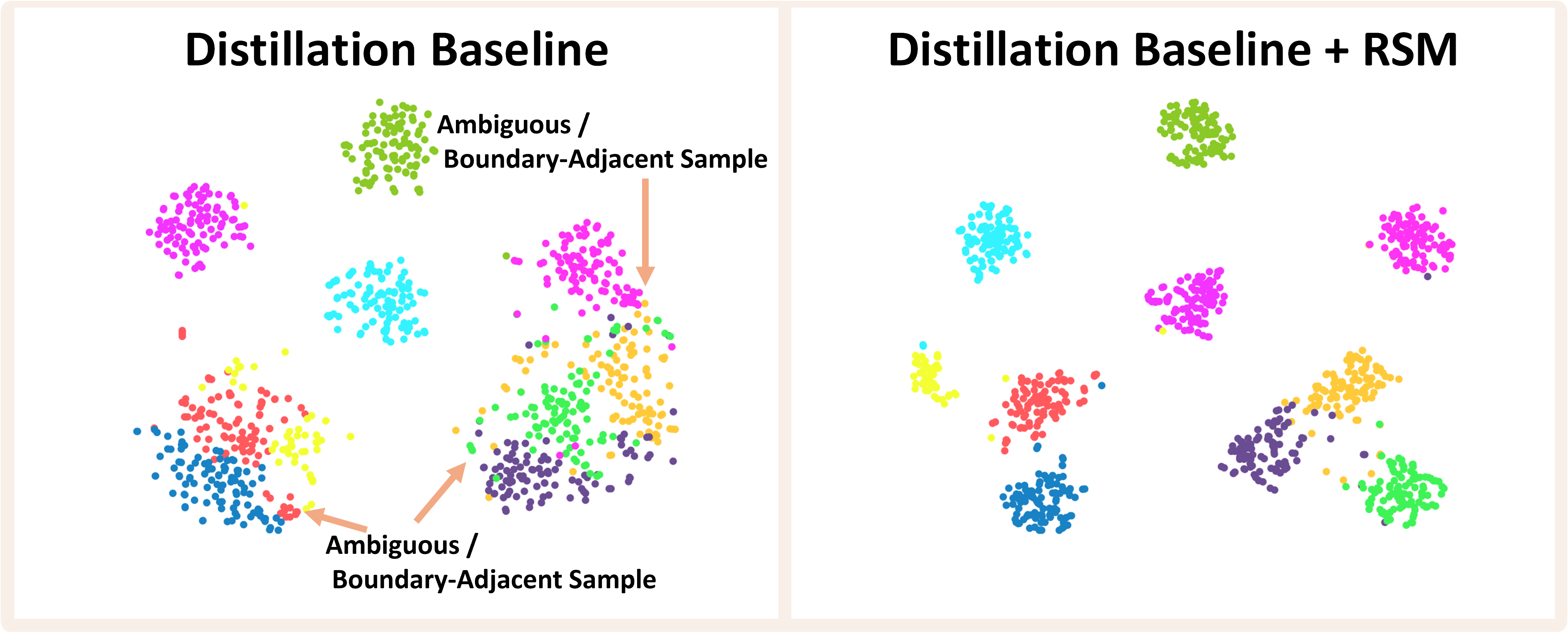}
    \subcaption{OxfordPets embedding visualization under 16-shot setting.}
  \end{minipage}

  \begin{minipage}[t]{0.86\linewidth}
    \centering
    \includegraphics[width=\linewidth]{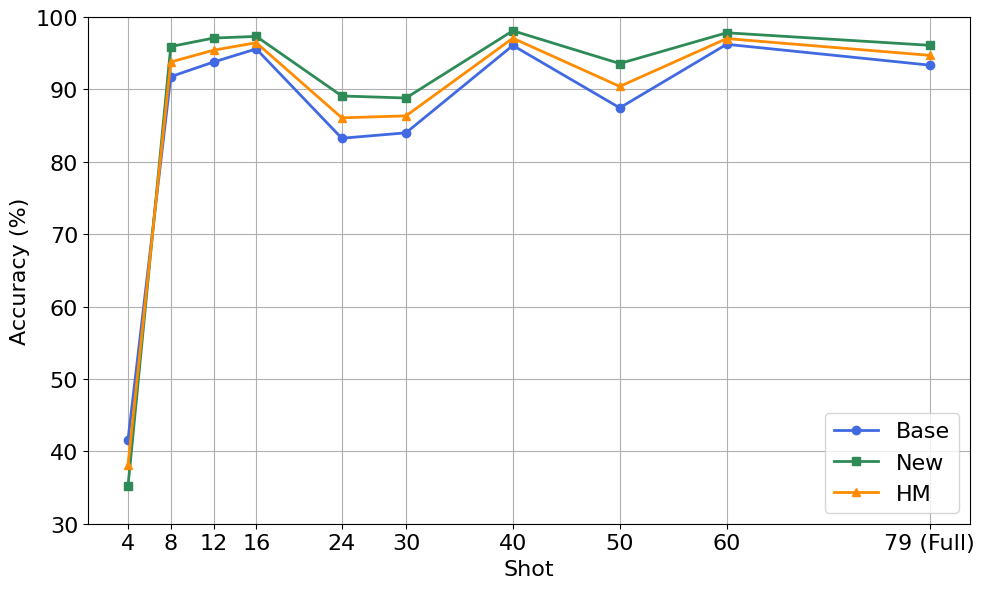}
    \subcaption{Performance on the OxfordPets across varying shots.}
  \end{minipage}
  \captionsetup{labelsep=period}
\captionsetup{justification=raggedright}

  \caption{Overview of representation quality and performance stability on OxfordPets under few-shot settings.
(a) highlighting inter-class boundaries and representation separability across methods.
shows the performance fluctuations across different shot settings over three evaluation metrics, which are detailed in Sec.~\ref{evaproto}.}
  \label{three_images}
\end{figure}

In essence, knowledge distillation transfers knowledge from powerful pretrained models to lightweight, task-specific ones, which makes it particularly well-suited for VLMs that benefit from extensive pretraining.
Recent cross-modal distillation methods, grounded in the CLIP architecture, have explored various strategies to transfer knowledge across modalities.
Notable examples include PromptKD~\cite{li2024promptkd} and KDPL~\cite{mistretta2024improving}, which primarily focus on distilling the knowledge encoded in learnable prompts to improve downstream performance. 
Although both approaches are effective, PromptKD requires full-shot training to achieve superior performance, while KDPL yields only marginal improvement in generalization.

From the standpoint of generalization, optimal performance in cross-modal distillation typically requires sufficient training data.
This dependency arises from the difficulty in generalizing ambiguous or boundary-adjacent representations, as illustrated in Figure~\ref{three_images}a (left), which often leads to high predictive uncertainty.
Otherwise, the performance curve illustrated in Figure~\ref{three_images}b tends to fluctuate in the presence of unfavorable geometric samples, until the training set becomes sufficiently large to enable generalization over such samples.
This observation is consistent with the finding in~\cite{phuong2019towards}, which highlights the role of data geometry, particularly class separation, in determining the success of knowledge distillation.
Yet in real-world scenarios, access to large-scale labeled data is often impractical, making it difficult to support such task-specific distillation training.
Thus, the major challenge in vision-language distillation training lies in achieving robust generalization under data-scarce conditions, particularly against unfavorable data geometry, such as samples with ambiguous or boundary-adjacent representation.

Addressing the coupled challenges of uncertainty and feature representation, information entropy stands as a theoretically grounded and effective mechanism that has demonstrated success across a broad spectrum of representation learning approaches.
However, our experiments reveal that directly compressing the information entropy over separate text and image feature spaces does not guarantee effective training, potentially due to misaligned semantic representations across modalities.
A principled approach, grounded in the manifold hypothesis~\cite{bengio2013representation}, is to embed multi-modal features into a shared low-dimensional manifold.
By enforcing structural alignment across modalities, the resulting representations enable more effective entropy minimization over the shared manifold, thereby enhancing the performance of knowledge distillation.

To this end, we propose \textbf{A}ligned \textbf{M}anifold \textbf{E}ntropy for Robust Vision-Language Distillation (AME), which reconfigures a shared cross-modal manifold to enable effective entropy minimization.
Consequently, AME compresses the cross-modal feature representations within the reconfigured shared manifold, particularly for ambiguous or boundary-adjacent samples, thereby facilitating robust knowledge distillation under low-data regimes.
The proposed mechanism is initialized as a lightweight, plug-and-play module, termed \textbf{R}econfigured \textbf{S}hared \textbf{M}anifold compression (RSM), that integrates seamlessly into existing distillation pipelines. 
Notably, it requires no architectural modifications to the backbone, ensuring broad compatibility to the vision-language distillation frameworks.

Specifically, we draw inspiration from Maximum Ratio Combining (MRC) in multi-antenna systems~\cite{simon2004digital}, where the signals received from multiple antennas are coherently combined by multiplying each with the complex conjugate of its corresponding channel coefficient, followed by summation.
Analogously, we treat the two modality features as signals transmitted from distinct antennas and learn a pair of projection functions that act as complex-conjugate counterparts.
This formulation facilitates the consistent projection of both modalities into a shared manifold, while the perspective inspired by MRC offers a principled foundation for learning structural cross-modal representations as illustrated in Figure~\ref{three_images}a (right).

As a result, our proposed AME demonstrates average generalization performance across 11 datasets under the few-shot setting. In specific, AME achieves significant improvements to the base distillation model for Base-to-New generalization and Cross-Dataset generalization, with performance gains of at least 2.79\% and 1.92\%, respectively.
Notably, our theoretical analysis demonstrates that the joint effect of distillation training and entropy minimization leads to a tighter generalization error bound.
This theoretical insight aligns with our empirical results, offering a principled guarantee for robust knowledge distillation in real-world scenarios.

In summary, this paper makes the major contributions as follows:
\begin{itemize}
    \item We propose a vision-language knowledge distillation model, Aligned Manifold Entropy for Robust Vision-Language Distillation (AME), which leverages entropy minimization over a reconfigured shared manifold of multi-modal features, enabling superior generalization under low-data regimes.
    
    \item We propose Reconfigured Shared Manifold compression (RSM) is a plug-and-play regularization module that can be seamlessly adopted in existing vision-language knowledge distillation methods to improve their performance further.

    \item We theoretically derive a generalization error bound that explains how the synergy between distillation and entropy minimization contributes to improved generalization.
    
    \item Extensive experiments on different settings (distillation architecture and training setting) suggest that proposed AME significantly improves distillation baselines in generalization capability across a wide range of downstream tasks. 
\end{itemize}

\section{Literature Review}
\subsection{Vision-Languange Models}
Vision language models (VLMs) have made significant advances in learning multi-modal representations through training on large-scale datasets. 
These representations exhibit strong generalization across a variety of downstream tasks, including image recognition ~\cite{recognition1,recognition2,recognition3}, object detection~\cite{object-detection1,object-detection2,object-detection3,object-detection4}, and segmentation~\cite{segmentation1,segmentation2,segmentation4}. 
Building on this foundation, several approaches, such as BAN~\cite{BAN}, Intra-Inter~\cite{Intra-Inter}, and MCAN~\cite{MCAN}, improve task performance by leveraging attention-based architectures. 
In contrast, models like ViLBERT~\cite{ViLBERT}, LXMERT~\cite{LXMERT}, and UNITER~\cite{UNITER} focus on vision-language learning through BERT-style architectures, achieving further performance gains.
Among these approaches, CLIP~\cite{CLIP}, ALIGN~\cite{ALIGN}, and LiT~\cite{zhai2022lit} have catalyzed a shift toward a new paradigm that aligns images and texts via their respective encoders, leading to superior performance in downstream tasks. 
However, compared to fine-tuned methods~\cite{dong2022clip}, pretrained VLMs exhibit suboptimal performance on specific tasks. This has motivated a line of research aimed at efficiently adapting vision-language models (VLMs) or transferring their knowledge to models tailored for specific downstream applications~\cite{gu2021open,luddecke2022image,wang2023improving}.

\begin{figure*}[htbp]

  \includegraphics[width=0.80\linewidth]{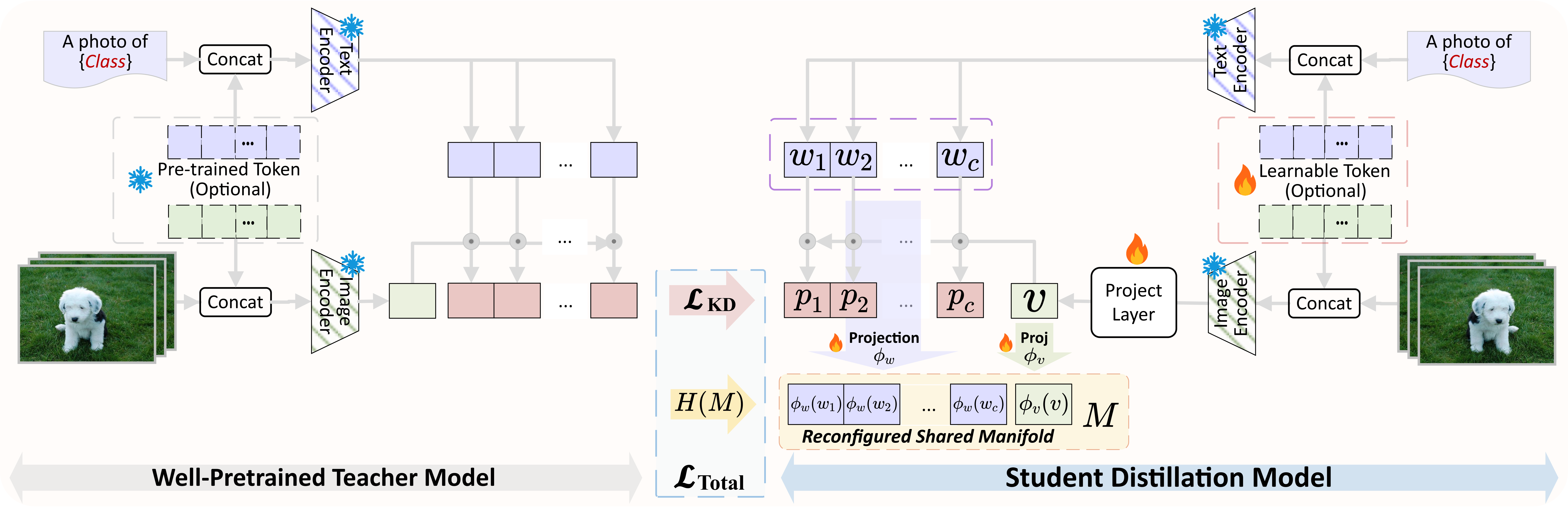}
  \captionsetup{labelsep=period}
\captionsetup{justification=raggedright}
\centering
  \caption{{Overview of Aligned Manifold Entropy for Robust Vision-Language Distillation (AME).} Given a well-pretrained teacher model and a lightweight student architecture, AME reconfigures a shared manifold for cross-modal feature compression via entropy minimization. The reconfigured shared manifold functions as a plug-and-play module that can be seamlessly integrated into a wide range of vision-language models, enabling structurally aligned cross-modal representation.}
  \label{main}
\end{figure*}

\subsection{Knowledge Distillation}
Knowledge distillation (KD)~\cite{hinton2015distilling} is a widely adopted technique in machine learning, in which a lightweight model (student) is trained to match output logits or intermediate representations of a larger pretrained model (teacher). 
Consequently, the student model can achieve improved task-specific performance, such as image classification~\cite{beyer2022knowledge, chen2019knowledge, hinton2015distilling} and image segmentation~\cite{dou2020unpaired,he2019knowledge}, while maintaining a compact architectural design.
Among the proposed KD methods, DML~\cite{zhang2018deep} introduces a strategy to train the student and teacher models simultaneously, thereby improving distillation performance without relying on a powerful teacher network.
Alternatively, DKD~\cite{zhao2022decoupled} decouples the distillation loss into two components, target-class KD and non-target-class KD, to better characterize the difficulty of training samples and explain the effectiveness of logit-based distillation. 
PromptKD~\cite{li2024promptkd} extends knowledge distillation to the prompt tuning paradigm, demonstrating superior generalization across a variety of downstream tasks.
More recently, KDPL~\cite{mistretta2024improving} improves distillation efficiency by updating only a limited number of parameters (i.e., the prompt), making the process more parameter-efficient.
In contrast to existing baselines~\cite{yun2020regularizing,yuan2020revisiting,zhang2018deep,zhao2022decoupled,li2024promptkd,mistretta2024improving}, we propose an entropy-based method that regularizes the output of the student model, aiming to enhance distillation performance under low-data regimes.


\section{Methodology}
Since entropy provides a principled measure of predictive uncertainty, incorporating it offers a natural means of enhancing representation effectiveness in knowledge distillation.
Building on this insight, we propose a method, namely \textbf{A}ligned \textbf{M}anifold \textbf{E}ntropy for Robust Vision-Language Distillation (AME), which redefines entropy minimization within a reconfigured shared manifold that integrates image and text embeddings.
This manifold serves to unify cross-modal feature representations, enabling the suppression of intra-class variability in the representation space, thereby achieving robust generalization in low-data regimes.
Prior to presenting AME, we first review the background of VLMs and knowledge distillation, which serve as our architectural foundation.

\subsection{Background}
\subsubsection{Vision Language Models. }
Existing vision language models, such as CLIP~\cite{CLIP}, are trained on large-scale image-text pair data, leaping forward the generalization capability across a wide range of multi-modal downstream tasks. 
Note that in CLIP, the image and text encoders transform the respective visual inputs and textual descriptions into embeddings, which are subsequently aligned using a contrastive loss. 
Specifically, the textual description is generated using the templates, e.g., ``a photo of a \{Class\}'',  where ${Class: C} \in \left \{1,2,...C\right \}$. 
Such descriptions are then fed into the text encoder $\mathcal{F}_w$, resulting in text embeddings $W = {\{w_i\}}_{i=1}^{C}$. 
The image embeddings $V = \{v_j\}_{j=1}^{N}$ are similarly generated by its encoder $\mathcal{F}_v$, where $N$ is the batch size. 
Finally, the predicted label $\hat{y}$ corresponds to the text embedding that achieves the highest cosine similarity $sim(\cdot)$ with the given image embedding, which can be formulated as
\begin{align}
    p(\hat{y}|v)= \frac{\mathrm{exp}\bigl( \mathrm{sim}(v, w_{\hat{y}})/\tau \bigr)}{\sum_{i=1}^{C} \mathrm{exp}\bigl( \mathrm{sim}(v, w_i)/\tau \bigr)},
\end{align}
where $\tau$ denotes a temperature parameter.

To unlock the potential of CLIP, a new paradigm, prompt learning has emerged to address the limitations of hand-crafted textual descriptions, which hinder the broader adoption of CLIP across downstream tasks. 
In such approaches, learnable tokens $P$ are prepended to the vectorized image and text inputs, which are then processed by the corresponding encoders, formally expressed as $V^p=\mathcal{F}_v([P^v, V])$ for the image branch and $W^p=\mathcal{F}_w([P^w, W])$ for the text branch, where $P^v$ and $P^w$ denote the visual and textual prompts, respectively.
Accordingly, the cosine similarity score used to compute the prediction probability is redefined as
\begin{align}
    p(\hat{y}|v^p)= \frac{\mathrm{exp}\bigl( \mathrm{sim}(v^p, w_{\hat{y}}^p)/\tau \bigr)}{\sum_{i=1}^{C} \mathrm{exp}\bigl( \mathrm{sim}(v^p, w_i^p)/\tau \bigr)}.
\end{align}

\subsubsection{Knowledge Distillation. }
In knowledge distillation (KD), to replicate the predictive behavior of a well-trained teacher model by minimizing the discrepancy between their outputs (e.g., logits or soft labels). Here, the Kullback-Leibler (KL) divergence loss is commonly used to calculate the discrepancy, and is defined as
\begin{align}
    \mathcal{L}_{KD} = \tau^2 D_{KL}(\sigma(z_t / \tau) \parallel \sigma(z_s / \tau)).
\end{align}
Here, $\sigma$ denotes the softmax function and $\tau$ is the temperature parameter, while $z_t$ and $z_s$ represent the prediction logits of the well-trained teacher and student models, respectively.

\subsection{Aligned Manifold Entropy for Robust Vision-Language Distillation (AME)}
Existing knowledge distillation frameworks for vision-language models (VLMs) often rely on large-scale training data to achieve reliable generalization, particularly for samples with ambiguous or boundary-adjacent representations. This reliance limits their applicability to downstream tasks in real-world scenarios where labeled data is scarce.
To this end, we propose Aligned Manifold Entropy for Robust Vision-Language Distillation (AME), a general approach that minimizes the entropy of a reconfigured shared manifold comprising both image and text features, thereby improving the effectiveness of knowledge distillation under limited data. The overall architecture of the proposed AME is illustrated in Figure~\ref{main}.
Specifically, AME enables structural cross-modal feature compression, which encourages well-separated representations and thereby achieve robust generalization under low-data regimes.
Notably, this lightweight mechanism is applied only during training and introduces no additional cost in the inference process.

\subsubsection{Reconfigured Shared Manifold}
Following the Manifold Hypothesis~\cite{tenenbaum2000global,roweis2000nonlinear}, high-dimensional data such as images and texts are assumed to reside on distinct low-dimensional manifolds within their respective embedding spaces. 
To facilitate intra-class determinacy in cross-modal feature representation, it is essential to jointly project both modalities into a shared manifold that enables structural feature compression.
However, merely projecting features into a joint space without additional constraints may fail to ensure semantic cohesion, potentially resulting in poorly aligned and high-entropy representations.
Moreover, in such approaches, the embedding representations within the manifold are primarily derived from the backbone of the distillation model, which not only limits the expressiveness of the manifold representations due to architectural constraints but also undermines the overall effectiveness of knowledge distillation.
To address this limitation, we introduce a learnable projection function that enables the shared manifold to capture and adapt to the intrinsic information of each modality.

Accordingly, the image and text embeddings, processed by their respective encoders, are projected via a convolution-based function $V' = \phi_{v}(V)$ and an MLP-based function $W' = \phi_{w}(W)$.
These resulting representations are used exclusively within the reconfigured shared manifold.
Subsequently, the reconfigured latent space $M \in \mathbb{R}^{(N+C) \times R}$ is formed by row-wise concatenation of the projected text embeddings and image embeddings, such that $M = \begin{bmatrix} \phi_w(w_{i=1}), \dots, \phi_w(w_{i=C}), \phi_v(v_{j=1}), \dots, \phi_v(v_{j=N}) \end{bmatrix}^{\top}$.

The resulting shared manifold is averaged along the feature dimension to obtain scalar representations, formulated as follows:
\begin{align}
    s_t = \frac{1}{R} \sum_{r=1}^{R} M_{t,r}, \quad t = 1, \dots, N+C.
\end{align}

A softmax operation is subsequently applied to the scalar scores $s_t$ to obtain the normalized probabilities $p_t$, representing the relative importance of each embedding within the reconfigured shared manifold $M$, which integrates both text and image representations. The information entropy of the manifold $M$ is then defined as:
\begin{equation}
 H(M) = - \sum_{t}^{N+C} p_t \log p_t,
\end{equation}
Accordingly, the total loss function, incorporating information entropy minimization within the manifold $M$, is defined as:
\begin{equation}
\mathcal{L}_{\text{Total}} = \mathcal{L}_{KD} + \omega {H}(M),
\label{total}
\end{equation}
where $\omega$ is a weighting coefficient, which is set to 50 in our case .
\subsection{Theoretical Analysis}
To gain deeper insight into the underlying mechanisms of the proposed AME framework, we conduct a theoretical analysis grounded in information theory and generalization theory. Specifically, this section formalizes the role of entropy minimization and derives an upper bound on the generalization error under the proposed distillation paradigm. The proofs are provided in the Appendix.

\begin{figure}[htbp]

\centering
  \includegraphics[width=0.86\linewidth]{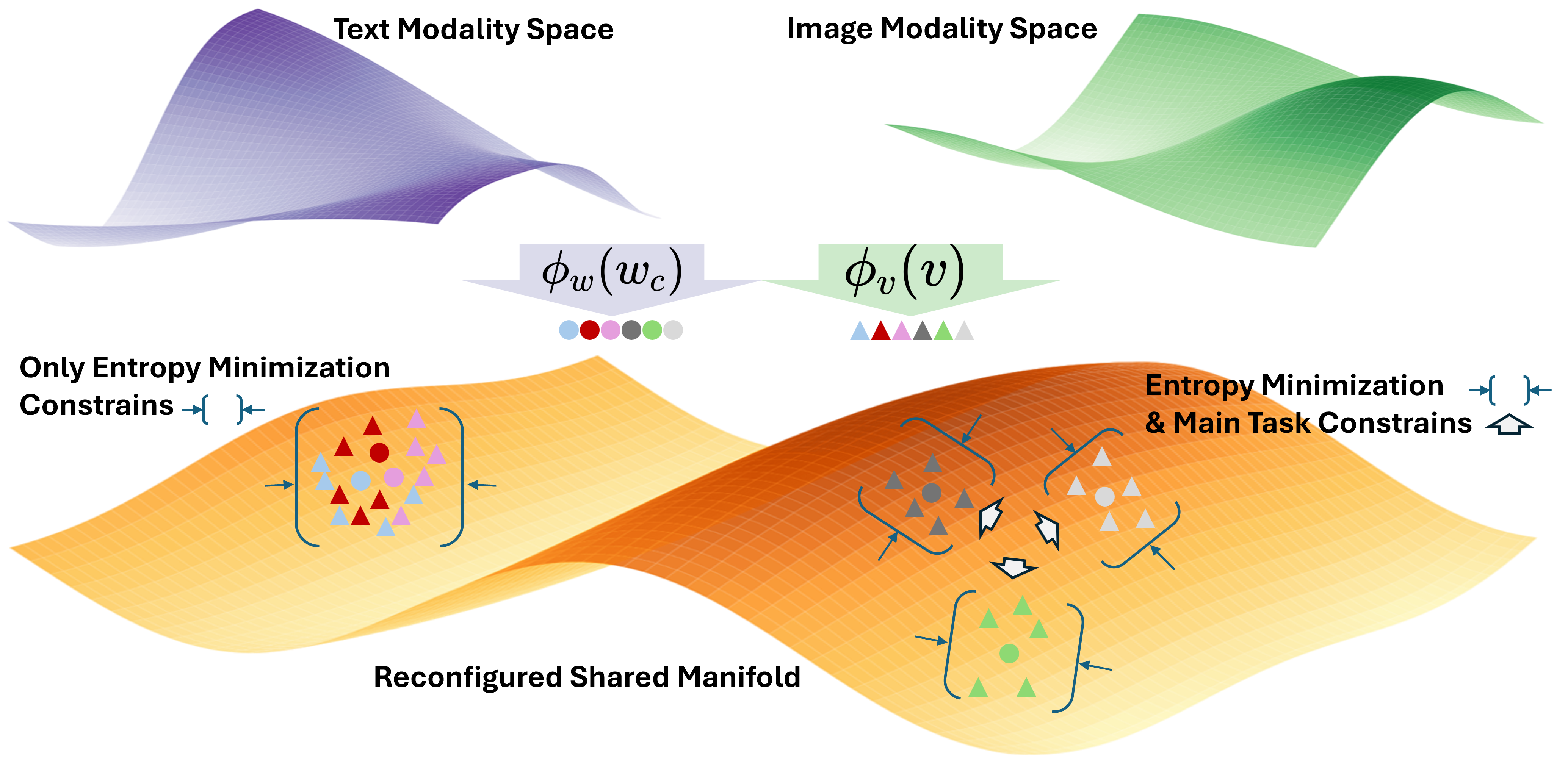}
    \captionsetup{labelsep=period }
\captionsetup{justification=raggedright}
  \caption{Illustration of modality spaces and reconfigured shared manifold. The upper part depicts the initial text modality space $\phi_w(W)$ (left) and image modality space $\phi_v(V)$ (right). The lower part shows the reconfigured shared manifold where both image and text features are projected and jointly optimized. On the left, only entropy minimization is applied, leading to less aligned or ambiguous clusters. On the right, the joint supervision of entropy minimization and main task constraints facilitates tighter and more structurally aligned representations.}
  \label{x}
\end{figure}

By characterizing the interactions between the reconfigured shared manifold and the optimization objective, we aim to explore the mechanism through which the proposed method enhances performance. 
Under typical optimization dynamics, minimizing $H(M)$ encourages each probability $p_t$ to approach a one-hot distribution. This implies that the corresponding score $s_t$ must be maximized at a specific index to effectively reduce entropy. Accordingly, the gradient of the loss with respect to $s_t$ can be expressed as:

\begin{align}
\frac{\partial H(M)}{\partial s_t} = - (1 + \log p_t).
\end{align}
This gradient behavior encourages all $s_t$ values to converge toward the dominant score $s_{t^*}$, leading the corresponding feature vectors $M_{t,:}$ vectors to collapse toward a single point in the reconfigured shared manifold. As a result, the logits in $\mathcal{L}_{\text{KD}}$ become degenerate, ultimately undermining the class separability necessary for effective knowledge distillation, as illustrated in the lower left of Figure~\ref{x}.

\subsubsection{Implicit class-wise alignment in reconfigured shared manifold}
\begin{theorem}
\label{Implicit Class-wise}
Given the task constraint imposed by $\mathcal{L}_{KD} $, jointly minimizing the total loss $\mathcal{L}_{\text{Total}}$ requires the optimal solution to achieve intra-class concentration while preserving inter-class separability.
At convergence, the optimization ensures that for any text embedding $w_c$ and its corresponding set of image embeddings $\{v_j: \text{label}(j)=c\}$ belonging to class $c$, their representations in the reconfigured shared manifold $M$ satisfy the following condition:
\begin{align}
\phi_w(w_c) = \phi_v(v_j) = h_c, \quad &\forall j: \text{label}(j)=c,\\
    \|h_c - h_{c'}\| \geq \zeta > 0, \quad &\forall c \neq c'.
\end{align}
Consequently, the embeddings belonging to the same class collapse toward a class-specific representation vector $h_c$, effectively reducing intra-class entropy, as illustrated in the lower right of Figure~\ref{x}. Meanwhile, embeddings from different classes maintain sufficient separation to satisfy the knowledge distillation constraint $\mathcal{L}_{KD}$, achieving structured compression of cross-modal representations.
\end{theorem}

The theorem demonstrates that the KL divergence loss between the teacher and student models enforces probabilistic compression between text and image. This loss implicitly induces a conditional joint distribution $p(W, V \mid S)$ over the reconfigured shared manifold, where $S$ denotes the input sample (e.g., an image-text pair). As a result, the information entropy $H(M)$ can be interpreted as an approximation of the conditional joint entropy $H'(W, V \mid S)$.

\subsubsection{Influence of the information entropy term on mutual information between the modalities, i.e., text $W$ and image $V$.}
\begin{corollary} 
By incorporating $H(M)$ into the loss function, minimizing $H(M)$, which approximates the conditional joint entropy $H(W, V \mid S)$), effectively increases the conditional mutual information $I(W; V \mid S)$ beyond what is achieved by the KL divergence term alone.
\label{co}
\end{corollary}

This reveals a dual mechanism: probabilistic alignment via KL divergence and structural compression via entropy minimization, both of which jointly improve cross-modal representation learning.

\subsubsection{Generalization bound via entropy regularization. }
\begin{corollary}
\label{mutual InB}
By incorporating entropy minimization over the reconfigured shared manifold $M$, which encodes the projected features of the student model, into the distillation framework, the mutual information between the model parameters and the training data is implicitly constrained. This leads to the generalization bound:
\begin{align}
\mathbb{E}_{\mathcal{S}, \theta} \bigl[ L(\theta; \mathcal{D}) - L(\theta; \mathcal{S}) \bigr]
\;\lesssim\;
\sqrt{\frac{2 (\delta + \epsilon)}{n}},
\end{align}
where $\theta$ denotes the parameters of the student model, $\mathcal{D}$ and $\mathcal{S}$ represent the training and test datasets respectively, and $n$ is the number of training samples. The term $\delta$ quantifies the entropy $H(M)$ of the reconfigured shared manifold $M$, while $\epsilon$ captures residual dependencies that are not eliminated through the information entropy $H(M)$.
\end{corollary}

This corollary demonstrates that the reconfigured shared manifold achieves an effective balance between compression and information preservation. Specifically, the entropy term $\delta = H(M)$ is explicitly minimized via the loss function, while the residual mutual information $\epsilon = I(\theta;\mathcal{S} \mid M)$ is implicitly suppressed, facilitated by the inclusion of a learnable projection function.
In other words, the combined quantity $\delta + \epsilon$ is strictly smaller than the original mutual information $I(\theta; \mathcal{S})$ under the KL-only setting. This reduction tightens the generalization bound, thereby highlighting the effectiveness of entropy minimization within the reconfigured shared manifold.


\subsubsection{Effect of Sample Size.}
As shown in Corollary~\ref{mutual InB}, the generalization bound decreases at a rate inversely proportional to the square root of the sample size $n$. Formally, it can be expressed as:
\begin{align}
\mathcal{O} \biggl( \sqrt{\tfrac{\delta + \epsilon}{n}} \biggr).
\tag{1}
\end{align}
When the sample size $n$ is large, the generalization gap naturally diminishes, and the contribution of entropy regularization becomes marginal, as data sufficiency dominates the learning dynamics.
In contrast, under low-data regimes, the entropy term $\mathcal{H}(M)$ becomes crucial. By minimizing $\delta$, it compensates for the effect of limited data and helps maintain a low generalization error.
This observation highlights the distinctive advantage of entropy regularization in cross-modal knowledge distillation under low-data regimes.

\begin{table*}[htbp]

\captionsetup{labelsep=period}
\captionsetup{justification=raggedright,singlelinecheck=false}
\captionof{table}{Comparison with existing approaches across 11 datasets and their averages. Here, RSM modules are appended to two distillation architectures, Simple and PromptKD, and are denoted by “+ RSM”. While \(\dagger\) indicates 16-shot training and \(\ddagger\) indicates full-shot training, with the number of shots shown in parentheses after the dataset name.}

\label{bigtable}
\captionsetup{justification=centering}

\begin{minipage}[t]{0.3\textwidth}
\centering
\makeatletter\def\@captype{table}
\captionsetup{skip=1.5pt}
\caption*{{(a) Average over 11 datasets}}
\centering
\setlength{\tabcolsep}{2.6pt}
\resizebox{0.875\textwidth}{!}{%
\begin{tabular}{lccc}
    \toprule
    Name     & Base& New & HM\\
    \midrule
    CLIP& 69.34& 74.22& \multicolumn{1}{|c}{71.70}   \\
    CoOp& {82.69}& 63.22& \multicolumn{1}{|c}{71.66
} \\
    MaPLe&  {82.28}& {75.14}& \multicolumn{1}{|c}{{78.55}} \\  
PromptSRC & 84.19 & 75.28 & \multicolumn{1}{|c}{79.49} \\
PromptKD\(^\dagger\) & 80.36 & 74.64 & \multicolumn{1}{|c}{77.40} \\
PromptKD\(^\ddagger\) & 85.21 & 79.13 & \multicolumn{1}{|c}{82.06} \\
\midrule
Simple + RSM\(^\dagger\) & 82.57 & 77.18 & \multicolumn{1}{|c}{79.78} \\
Simple + RSM\(^\ddagger\) & 85.08 & 79.21 & \multicolumn{1}{|c}{82.04} \\
PromptKD + RSM\(^\dagger\) & 83.31 & 77.30 & \multicolumn{1}{|c}{80.19} \\
PromptKD + RSM\(^\ddagger\) & 86.27 & 79.89 & \multicolumn{1}{|c}{82.96} \\

    \bottomrule
\end{tabular}
}
\end{minipage}
\hspace{0.1in}
\begin{minipage}[t]{0.3\textwidth}
\makeatletter\def\@captype{table}
\captionsetup{skip=1.5pt}
\caption*{(b) ImageNet (\(\approx\) 1281)}
\centering
\setlength{\tabcolsep}{2.6pt}
\resizebox{0.875\textwidth}{!}{%
\begin{tabular}{lccc}
    \toprule
    Name     & Base& New & HM\\
    \midrule
    CLIP& 72.43& 68.14& \multicolumn{1}{|c}{70.22
}   \\
    CoOp&  {76.47}& 67.88& \multicolumn{1}{|c}{71.92
} \\
    MaPLe& {76.66}&  {70.54}& \multicolumn{1}{|c}{ {73.47}} \\   
PromptSRC & 77.36 & 70.58 & \multicolumn{1}{|c}{73.81} \\
PromptKD\(^\dagger\) & 77.02 & 71.03 & \multicolumn{1}{|c}{73.90} \\
PromptKD\(^\ddagger\) & 76.04 & 70.74 & \multicolumn{1}{|c}{73.30} \\
\midrule
Simple + RSM\(^\dagger\) & 74.53 & 68.84 & \multicolumn{1}{|c}{71.57} \\
Simple + RSM\(^\ddagger\) & 74.14 & 70.11 & \multicolumn{1}{|c}{72.07} \\
PromptKD + RSM\(^\dagger\) & 76.03 & 70.21 & \multicolumn{1}{|c}{73.00} \\
PromptKD + RSM\(^\ddagger\) & 75.73 & 71.04 & \multicolumn{1}{|c}{73.31} \\

    \bottomrule
\end{tabular}
}
\end{minipage}
\hspace{0.1in}
\begin{minipage}[t]{0.3\textwidth}
\makeatletter\def\@captype{table}
\captionsetup{skip=1.5pt}
\caption*{(c) Caltech101 (\(\approx\) 41)}
\centering
\setlength{\tabcolsep}{2.6pt}
\resizebox{0.875\textwidth}{!}{%
\begin{tabular}{lccc}
    \toprule
    Name     & Base& New & HM\\
    \midrule
    CLIP& 96.84& 94.00& \multicolumn{1}{|c}{95.40}   \\
    CoOp&  {98.00}& 89.81& \multicolumn{1}{|c}{93.73
} \\
    MaPLe& 97.74&  {94.36}& \multicolumn{1}{|c}{ {96.02}} \\   
PromptSRC & 97.98 & 93.92 & \multicolumn{1}{|c}{95.91} \\
PromptKD\(^\dagger\) & 98.34 & 95.23 & \multicolumn{1}{|c}{96.76} \\
PromptKD\(^\ddagger\) & 98.90 & 96.47 & \multicolumn{1}{|c}{97.67} \\
\midrule
Simple + RSM\(^\dagger\) & 97.89 & 95.31 & \multicolumn{1}{|c}{96.58} \\
Simple + RSM\(^\ddagger\) & 98.21 & 96.07 & \multicolumn{1}{|c}{97.13} \\
PromptKD + RSM\(^\dagger\) & 98.56 & 95.67 & \multicolumn{1}{|c}{97.09} \\
PromptKD + RSM\(^\ddagger\) & 98.82 & 96.07 & \multicolumn{1}{|c}{97.42} \\

    \bottomrule
\end{tabular}
}
\end{minipage}
\newline

\begin{minipage}[t]{0.3\textwidth}
\makeatletter\def\@captype{table}
\captionsetup{skip=1.5pt}
\caption*{(d) OxfordPets (\(\approx\) 79)}
\centering
\setlength{\tabcolsep}{2.6pt}
\resizebox{0.875\textwidth}{!}{%
\begin{tabular}{lccc}
    \toprule
    Name     & Base& New & HM\\
    \midrule
    CLIP& 91.17& 97.26& \multicolumn{1}{|c}{94.12
}   \\
    CoOp& 93.67& 95.29& \multicolumn{1}{|c}{94.47
} \\
    MaPLe& {95.43}& {97.76}& \multicolumn{1}{|c}{{96.58}} \\   
PromptSRC & 95.43 & 97.30 & \multicolumn{1}{|c}{96.35} \\
PromptKD\(^\dagger\) & 90.13 & 92.54 & \multicolumn{1}{|c}{91.32} \\
PromptKD\(^\ddagger\) & 93.39 & 95.79 & \multicolumn{1}{|c}{94.57} \\
\midrule
Simple + RSM\(^\dagger\) & 93.34 & 97.13 & \multicolumn{1}{|c}{95.20} \\
Simple + RSM\(^\ddagger\) & 95.46 & 98.17 & \multicolumn{1}{|c}{96.80} \\
PromptKD + RSM\(^\dagger\) & 95.23 & 97.07 & \multicolumn{1}{|c}{96.14} \\
PromptKD + RSM\(^\ddagger\) & 96.38 & 98.23 & \multicolumn{1}{|c}{97.30} \\
    \bottomrule
\end{tabular}
}
\end{minipage}
\hspace{0.1in}
\begin{minipage}[t]{0.3\textwidth}
\makeatletter\def\@captype{table}
\captionsetup{skip=1.5pt}
\caption*{(e) StanfordCars (\(\approx\) 32)}
\centering
\setlength{\tabcolsep}{2.6pt}
\resizebox{0.875\textwidth}{!}{%
\begin{tabular}{lccc}
    \toprule
    Name     & Base& New & HM\\
    \midrule
    CLIP& 63.37& {74.89}& \multicolumn{1}{|c}{68.65
}   \\
    CoOp& {78.12}& 60.40& \multicolumn{1}{|c}{68.13
} \\
    MaPLe&  {72.94}& 74.00& \multicolumn{1}{|c}{{73.47}} \\   
 PromptSRC & 78.53 & 75.23 & \multicolumn{1}{|c}{76.84} \\
PromptKD\(^\dagger\) & 79.48 & 81.40 & \multicolumn{1}{|c}{80.43} \\
PromptKD\(^\ddagger\) & 76.75 & 78.21 & \multicolumn{1}{|c}{77.47} \\
\midrule
Simple + RSM\(^\dagger\) & 78.30 & 79.79 & \multicolumn{1}{|c}{79.04} \\
Simple + RSM\(^\ddagger\) & 80.22 & 81.37 & \multicolumn{1}{|c}{80.79} \\
PromptKD + RSM\(^\dagger\) & 80.30 & 81.54 & \multicolumn{1}{|c}{80.92} \\
PromptKD + RSM\(^\ddagger\) & 82.60 & 83.44 & \multicolumn{1}{|c}{83.02} \\
    \bottomrule
\end{tabular}
}
\end{minipage}
\hspace{0.1in}
\begin{minipage}[t]{0.3\textwidth}
\makeatletter\def\@captype{table}
\captionsetup{skip=1.5pt}
\caption*{(f) Flowers102 (\(\approx\) 40)}
\centering
\setlength{\tabcolsep}{2.6pt}
\resizebox{0.875\textwidth}{!}{%
\begin{tabular}{lccc}
    \toprule
    Name     & Base& New & HM\\
    \midrule
    CLIP& 72.08& {77.80}& \multicolumn{1}{|c}{74.83
}   \\
    CoOp& {97.60}& 59.67& \multicolumn{1}{|c}{74.06
} \\
    MaPLe&  {95.92}& 72.46& \multicolumn{1}{|c}{ {82.56}} \\   

PromptSRC & 98.04 & 74.82 & \multicolumn{1}{|c}{84.87} \\
PromptKD\(^\dagger\) & 95.38 & 77.28 & \multicolumn{1}{|c}{85.38} \\
PromptKD\(^\ddagger\) & 96.26 & 80.05 & \multicolumn{1}{|c}{87.41} \\
\midrule
Simple + RSM\(^\dagger\) & 98.10 & 81.18 & \multicolumn{1}{|c}{88.84} \\
Simple + RSM\(^\ddagger\) & 99.02 & 81.87 & \multicolumn{1}{|c}{89.63} \\
PromptKD + RSM\(^\dagger\) & 98.58 & 81.65 & \multicolumn{1}{|c}{89.32} \\
PromptKD + RSM\(^\ddagger\) & 99.31 & 82.55 & \multicolumn{1}{|c}{90.16} \\
    \bottomrule
\end{tabular}
}
\end{minipage}
 \newline

\begin{minipage}[t]{0.3\textwidth}

\makeatletter\def\@captype{table}
\captionsetup{skip=1.5pt}
\caption*{(g) Food101 (\(=\) 500)}
\centering
\setlength{\tabcolsep}{2.6pt}
\resizebox{0.875\textwidth}{!}{%
\begin{tabular}{lccc}
    \toprule
    Name     & Base& New & HM\\
    \midrule
    CLIP& 90.10& 91.22& \multicolumn{1}{|c}{90.66
}   \\
    CoOp& 88.33& 82.26& \multicolumn{1}{|c}{85.19
} \\
    MaPLe&  {90.71}& {92.05}& \multicolumn{1}{|c}{{91.38}} \\   
PromptSRC & 90.84 & 91.54 & \multicolumn{1}{|c}{91.19} \\
PromptKD\(^\dagger\) & 78.56 & 81.29 & \multicolumn{1}{|c}{79.90} \\
PromptKD\(^\ddagger\) & 92.39 & 93.78 & \multicolumn{1}{|c}{93.08} \\
\midrule
Simple + RSM\(^\dagger\) & 88.48 & 90.51 & \multicolumn{1}{|c}{89.48} \\
Simple + RSM\(^\ddagger\) & 92.16 & 93.39 & \multicolumn{1}{|c}{92.77} \\
PromptKD + RSM\(^\dagger\) & 89.48 & 91.06 & \multicolumn{1}{|c}{90.26} \\
PromptKD + RSM\(^\ddagger\) & 92.52 & 93.73 & \multicolumn{1}{|c}{93.12} \\

    \bottomrule
\end{tabular}
}
\end{minipage}
\hspace{0.1in}
\begin{minipage}[t]{0.3\textwidth}
\makeatletter\def\@captype{table}
\captionsetup{skip=1.5pt}

\caption*{(h) FGVCAircraft (\(\approx\) 33)}
\centering
\setlength{\tabcolsep}{2.6pt}
\resizebox{0.875\textwidth}{!}{%
\begin{tabular}{lccc}
    \toprule
    Name     & Base& New & HM\\
    \midrule
    CLIP& 27.19& {36.29}& \multicolumn{1}{|c}{31.09
}   \\
    CoOp& {40.44}& 22.30& \multicolumn{1}{|c}{28.75
} \\
    MaPLe&  {37.44}&  {35.61}& \multicolumn{1}{|c}{{36.50}} \\   

PromptSRC & 42.82 & 36.77 & \multicolumn{1}{|c}{39.57} \\
PromptKD\(^\dagger\) & 44.42 & 39.81 & \multicolumn{1}{|c}{41.99} \\
PromptKD\(^\ddagger\) & 47.84 & 41.33 & \multicolumn{1}{|c}{44.35} \\
\midrule
Simple + RSM\(^\dagger\) & 43.68 & 37.27 & \multicolumn{1}{|c}{40.22} \\
Simple + RSM\(^\ddagger\) & 45.44 & 38.77 & \multicolumn{1}{|c}{41.84} \\
PromptKD + RSM\(^\dagger\) & 45.32 & 39.25 & \multicolumn{1}{|c}{42.07} \\
PromptKD + RSM\(^\ddagger\) & 48.34 & 41.39 & \multicolumn{1}{|c}{44.60} \\

    \bottomrule
\end{tabular}
}
\end{minipage}
\hspace{0.1in}
\begin{minipage}[t]{0.3\textwidth}
\makeatletter\def\@captype{table}
\captionsetup{skip=1.5pt}
\caption*{(i) SUN397 (\(\approx\) 40)}
\centering
\setlength{\tabcolsep}{2.6pt}
\resizebox{0.875\textwidth}{!}{%
\begin{tabular}{lccc}
    \toprule
    Name     & Base& New & HM\\
    \midrule
    CLIP& 69.36& 75.35& \multicolumn{1}{|c}{72.23
}   \\
    CoOp&  {80.60}& 65.89& \multicolumn{1}{|c}{72.51
} \\
    MaPLe& {80.82}& {78.70}& \multicolumn{1}{|c}{{79.75}} \\   

PromptSRC & 82.59 & 78.71 & \multicolumn{1}{|c}{80.60} \\
PromptKD\(^\dagger\) & 82.64 & 80.48 & \multicolumn{1}{|c}{81.55} \\
PromptKD\(^\ddagger\) & 83.48 & 81.38 & \multicolumn{1}{|c}{82.42} \\
\midrule
Simple + RSM\(^\dagger\) & 81.48 & 79.50 & \multicolumn{1}{|c}{80.48} \\
Simple + RSM\(^\ddagger\) & 82.81 & 80.84 & \multicolumn{1}{|c}{81.81} \\
PromptKD + RSM\(^\dagger\) & 82.25 & 79.94 & \multicolumn{1}{|c}{81.08} \\
PromptKD + RSM\(^\ddagger\) & 83.56 & 81.30 & \multicolumn{1}{|c}{82.41} \\
    \bottomrule
\end{tabular}
}
\end{minipage}
\newline

\begin{minipage}[t]{0.3\textwidth}
\makeatletter\def\@captype{table}
\captionsetup{skip=1.5pt}
\caption*{(j) DTD (\(=\) 60)}
\centering
\setlength{\tabcolsep}{2.6pt}
\resizebox{0.875\textwidth}{!}{%
\begin{tabular}{lccc}
    \toprule
    Name     & Base& New & HM\\
    \midrule
    CLIP& 53.24& {59.90}& \multicolumn{1}{|c}{56.37
}   \\
    CoOp& 79.44& 41.18& \multicolumn{1}{|c}{54.24} \\
    MaPLe& {80.36}&  {59.18}& \multicolumn{1}{|c}{{68.16}} \\   
PromptSRC & 82.83 & 61.19 & \multicolumn{1}{|c}{70.39} \\
PromptKD\(^\dagger\) & 75.42 & 60.51 & \multicolumn{1}{|c}{67.15} \\
PromptKD\(^\ddagger\) & 85.42 & 71.05 & \multicolumn{1}{|c}{77.58} \\
\midrule
Simple + RSM\(^\dagger\) & 81.94 & 68.16 & \multicolumn{1}{|c}{74.42} \\
Simple + RSM\(^\ddagger\) & 84.34 & 70.49 & \multicolumn{1}{|c}{76.80} \\
PromptKD + RSM\(^\dagger\) & 84.15 & 69.16 & \multicolumn{1}{|c}{75.92} \\
PromptKD + RSM\(^\ddagger\) & 86.04 & 70.37 & \multicolumn{1}{|c}{77.42} \\

    \bottomrule
\end{tabular}
}
\end{minipage}
\hspace{0.1in}
\begin{minipage}[t]{0.3\textwidth}
\makeatletter\def\@captype{table}
\captionsetup{skip=1.5pt}
\caption*{(k) EuroSAT (\(\approx\) 1350)}
\centering
\setlength{\tabcolsep}{2.6pt}
\resizebox{0.875\textwidth}{!}{%
\begin{tabular}{lccc}
    \toprule
    Name     & Base& New & HM\\
    \midrule
    CLIP& 56.48& 64.05& \multicolumn{1}{|c}{60.03
}   \\
    CoOp& 92.19& 54.74& \multicolumn{1}{|c}{68.69
} \\
    MaPLe& {94.07}&  {73.23}& \multicolumn{1}{|c}{ {82.35}} \\   

PromptSRC & 93.35 & 69.29 & \multicolumn{1}{|c}{79.54} \\
PromptKD\(^\dagger\) & 75.47 & 61.55 & \multicolumn{1}{|c}{67.80} \\
PromptKD\(^\ddagger\) & 97.32 & 80.27 & \multicolumn{1}{|c}{87.98} \\
\midrule
Simple + RSM\(^\dagger\) & 85.41 & 72.41 & \multicolumn{1}{|c}{78.38} \\
Simple + RSM\(^\ddagger\) & 96.58 & 79.92 & \multicolumn{1}{|c}{87.47} \\
PromptKD + RSM\(^\dagger\) & 79.94 & 65.03 & \multicolumn{1}{|c}{71.72} \\
PromptKD + RSM\(^\ddagger\) & 97.60 & 81.19 & \multicolumn{1}{|c}{88.64} \\
    \bottomrule
\end{tabular}
}
\end{minipage}
\hspace{0.1in}
\begin{minipage}[t]{0.3\textwidth}
\makeatletter\def\@captype{table}
\captionsetup{skip=1.5pt}
\caption*{(l) UCF101 (\(\approx\) 75)}
\centering
\setlength{\tabcolsep}{2.6pt}
\resizebox{0.875\textwidth}{!}{%
\begin{tabular}{lccc}
    \toprule
    Name     & Base& New & HM\\
    \midrule
    CLIP& 70.53& 77.50& \multicolumn{1}{|c}{73.85
}   \\
    CoOp& {84.69}& 56.05& \multicolumn{1}{|c}{67.46
} \\
    MaPLe& 83.00& {78.66}& \multicolumn{1}{|c}{{80.77}} \\   

PromptSRC & 86.35 & 78.71 & \multicolumn{1}{|c}{82.35} \\
PromptKD\(^\dagger\) & 87.12 & 79.90 & \multicolumn{1}{|c}{83.36} \\
PromptKD\(^\ddagger\) & 89.50 & 81.32 & \multicolumn{1}{|c}{85.22} \\
\midrule
Simple + RSM\(^\dagger\) & 85.07 & 78.85 & \multicolumn{1}{|c}{81.85} \\
Simple + RSM\(^\ddagger\) & 87.50 & 80.31 & \multicolumn{1}{|c}{83.75} \\
PromptKD + RSM\(^\dagger\) & 86.54 & 79.66 & \multicolumn{1}{|c}{82.96} \\
PromptKD + RSM\(^\ddagger\) & 88.09 & 79.52 & \multicolumn{1}{|c}{83.58} \\
    \bottomrule
\end{tabular}
}
\end{minipage}
\end{table*}

\begin{table*}[htbp]

\setlength{\tabcolsep}{3pt}
\captionsetup{labelsep=period}
\captionof{table}{Comparison on Cross-datasets. Here, RSM is appended to a simple distillation model and PromptKD. Methods marked with \(\dagger\) are trained under the 16-shot setting, while those marked with \(\ddagger\) are trained under the full-shot setting.}
\label{cross}
\resizebox{0.72\textwidth}{!}{%
\begin{tabular}{lcccccccccccc}
    \toprule

 Method    &  {Caltech}&  {Pets}&  {Cars}&  {Flowers}&  {Food}& {Aircraft} &  {SUN}&  {DTD}&  {EuroSAT}& {UCF} & {Avg.}\\
     \midrule
    CLIP  & 96.84 & 94.00 & 95.40 & 68.14 & 85.30 & 18.47 & 64.15 & 41.92 & 46.39 & 66.55 & 65.17 \\
    CoOp& 93.70& 89.14& 64.51& 68.71& 85.30& 18.47& 64.15& 41.92& 46.39 & 66.55&63.88
\\
    MaPLe& 93.53& {90.49}&  {65.57}& {72.23}& {86.20}& {24.74}& 67.01& {46.49}&  {48.06}&  {68.69}& {66.30}\\
    PromptSRC & 93.71 & 90.39 & 65.51 & 70.37 & 86.37 & 23.27 & 67.36 & 46.24 & 43.78 & 68.44 & 65.54 \\
    PromptKD$^{\dagger}$ & 93.40 & 75.03 & 71.25 & 73.04 & 84.63 & 24.84 & 67.50 & 47.34 & 38.35 & 72.39 & 64.78 \\
PromptKD$^{\ddagger}$ & 88.95 & 91.96 & 73.99 & 75.11 & 88.92 & 25.97 & 68.52 & 55.52 & 62.00 & 75.63 & 70.66 \\
    \midrule
Simple+ RSM$^{\dagger}$ & 93.17 & 87.31 & 69.68 & 73.27 & 83.86 & 24.26 & 66.00 & 50.95 & 48.72 & 71.42 & 66.86 \\
Simple+ RSM$^{\ddagger}$ & 88.71 & 85.30 & 61.82 & 70.04 & 85.63 & 20.65 & 62.72 & 49.90 & 52.42 & 69.19 & 64.64 \\
PromptKD + RSM$^{\dagger}$ & 93.74 & 87.96 & 70.84 & 73.94 & 84.73 & 24.46 & 66.77 & 52.52 & 39.75 & 72.27 & 66.70 \\
PromptKD + RSM$^{\ddagger}$ & 93.13 & 91.57 & 73.25 & 75.13 & 88.95 & 25.90 & 68.23 & 55.24 & 60.79 & 75.36 & 70.76 \\
\bottomrule

\end{tabular}
}
\end{table*}

\section{Experiments}
AME is a general method requiring no architectural modifications to the backbone. It can therefore be deployed as a plug-and-play module, referred to as RSM, allowing seamless integration into a wide range of knowledge distillation frameworks for performance enhancement under data-scarce conditions.
To assess the effectiveness of RSM, we apply it to both PromptKD and a simplified architecture, referred to as Simple, where the main CLIP backbone in Simple is frozen and a learnable projection function is added to the image branch after the encoder.
Following the evaluation protocol of PromptKD, we compare these methods with well-established baselines across 11 datasets under two evaluations, i.e., Base-to-New Class and Cross-Dataset Generalization.
In this section, the details of the experimental setup, including the training and evaluation protocols, baselines, and datasets, are presented in sequence, followed by fundamental experiments and further analysis.


\subsection{Experiments Setup}

\subsubsection{Training and Evaluation protocols. } 
\label{evaproto}
The implementation details adhere to the settings as PromptKD unless otherwise noted. 
Accordingly, all distillation methods in this paper use a well-trained PromptSRC~\cite{PROMPTSRC_Khattak_2023_ICCV} as teacher model, which is provided by PromptKD. 
For the training of the student model, CLIP with the ViT-B/16 backbone is used, while optimization is performed for 20 epochs with the stochastic gradient descent (SGD) optimizer, using a batch size of 8 and a learning rate of 0.005.
The evaluation metric is defined as the average accuracy over three runs using random seeds 1, 2, and 3. For each run, accuracies are computed separately for base classes (Base), new classes (New), and their harmonic mean (HM). Then the final results are obtained by averaging these values across runs.

\subsubsection{Baselines. } 
In addition to PromptKD, we consider several well-known prompt tuning approaches as baselines for our evaluation, including CLIP, CoOp, MaPLe, and PromptSRC, which cover both single-modality and multi-modality prompt tuning paradigms.

\subsubsection{Datasets. }
To ensure consistency with prior works~\cite{COOP,COCOOP,khattak2023maple,recognition1,li2024promptkd}, we adopt the evaluation protocol on the same datasets.
For the Base-to-New Class Generalization, eleven image datasets are used. 
In the cross-dataset generalization setting, we follow the protocol of PromptKD, evaluating on ten datasets with ImageNet excluded.
Specifically, Caltech101~\cite{Caltech101} and ImageNet~\cite{ImageNet} are categorized as generic-object datasets; 
FGVCAircraft~\cite{FGVCAircraft}, Flowers102~\cite{Flowers102}, Food101~\cite{Food101}, OxfordPets~\cite{OxfordPets} and StanfordCars~\cite{StanfordCars} are fine-grained image datasets; 
DTD~\cite{DTD} and EuroSAT~\cite{EuroSAT} are texture and satellite datasets, respectively, while SUN397~\cite{SUN397} and UCF101~\cite{UCF101} are used for scene and action recognition tasks, respectively. 

\subsection{Fundamental Experiments}
\subsubsection{Base-to-New Class Generalization. }
Following~\cite{li2024promptkd}, both the training and test datasets are split into base and new classes. The student model is trained on the unlabeled training set, while its performance is evaluated on the test set for both base and new classes. 
As shown in Table~\ref{bigtable}, we evaluate whether integrating our proposed RSM module, denoted as “+ RSM”, into existing KD models improves knowledge transfer under both 16-shot and full-shot settings, by comparing with the baselines.
Note that in the full-shot setting, all KD models, including PromptKD, Simple + RSM, and PromptKD + RSM, are trained on ImageNet for two epochs.

Under the 16-shot training setting for distillation, it can be calculated in Table~\ref{bigtable} that PromptSRC achieves the best average performance among all baselines across 11 datasets, achieving 84.19\%, 75.28\% and 79.49\% for Base,  New, and HM, respectively.  
Alternatively, PromptKD exhibits suboptimal performance, with corresponding results of 80.36\%, 74.64\% and 77.40\%.
In contrast, by integrating our proposed module, Simple + RSM achieves competitive results of 82.57\%, 77.18\% and 79.78\%, while PromptKD + RSM further improves upon PromptKD, reaching 83.31\%, 77.30\% and 80.19\%, on Base, New, and HM, respectively.
Notably, PromptKD + RSM yields significant improvements over PromptKD by 2.94\%, 2.66\% and 2.79\% on Base, New, and HM, respectively.
It can be clearly seen that both model (i.e., Simple + RSM and PromptKD + RSM) surpass the best generalization provide by baselines on both the New class and HM metrics, and achieve the best HM on 7 out of 11 datasets, including Caltech101, StanfordCars, Flowers102, FGVCAircraft, SUN397 DTD and UCF101.  
Note that on OxfordPets, Flowers102, Food101, DTD, and EuroSAT, PromptKD + RSM outperforms PromptKD in terms of HM, achieving 96.14\%, 89.32\%, 90.26\%, 75.92\% and 71.72\%, respectively. These correspond to relative improvements of 4.82\%, 3.94\%, 10.36\%, 8.78\% and 3.92\% over the original PromptKD.

With the full-shot training setting, it can be seen in Table~\ref{bigtable}, PromptKD obtains the best average performance for all evaluation metrics ( 85.21\%, 79.13\% and 82.06\% for Base, New, and HM, respectively) among the baselines.
By leveraging our proposed method, PromptKD + RSM achieves further improvement in overall performance, with gains of 1.06\%, 0.77\% and 0.90\% on the Base, New, and HM, respectively. In terms of HM, improvements are observed on 7 out of 11 datasets including ImageNet, OxfordPets, StanfordCars, Flowers102, Food101, FGVCAircraft, and EuroSAT.
Notably, RSM significantly improves the performance of both Simple-based and PromptKD-based distillation models across all metrics on OxfordPets, StanfordCars, and Flowers102. For example, in terms of HM, the improvements from the PromptKD + RSM are 2.73\%, 5.54\%, and 2.75\% on OxfordPets, StanfordCars, and Flowers102, respectively.

Overall, based on the comparison across different foundation models and the training setting in Table~\ref{bigtable}, our proposed RSM is shown to consistently enhance generalization performance in knowledge distillation. 
These results suggest that our proposed RSM serves as a universal and effective module for vision-language knowledge distillation models, enabling robust generalization across a wide range of downstream tasks under the low-data regime.

\subsubsection{Cross-Dataset Generalization}
In Cross-Dataset generalization, the student model is trained on unlabeled images of unseen classes, similar to the Base-to-New Class generalization setting.
Table~\ref{cross} presents the performance comparison among the prompt tuning baselines, PromptKD, and the models augmented with our proposed RSM.
Specifically, the RSM-augmented model demonstrates superior generalization performance on 6 out of 10 datasets, as well as in the average performance across all datasets under the 16-shot setting.
Notably, under the 16-shot setting, RSM yields significant improvements in generalization performance when applied to PromptKD, with gains of 12.93\% on OxfordPets, 5.18\% on DTD, and 1.40\% on EuroSAT. These improvements result in an average gain of 1.92\% across all datasets.
Clearly, this analysis consistently demonstrates that the proposed RSM significantly improves the performance of vision-language knowledge distillation models.

\subsection{Further Analysis}
To further investigate the effectiveness and adaptability of our proposed RSM module, we conduct a series of extended analyses. Specifically, we integrate RSM into two well-established prompt tuning approaches, MaPLe and PromptSRC, to assess its compatibility across different architectures.
In addition, we evaluate its performance under different shot settings by comparing it with the original PromptKD.
We also analyze the role of the learnable projection used in the reconfigured shared manifold.
These investigations are presented and discussed in the following sections.

\subsubsection{Effectiveness of RSM for the Prompt Tuning paradigm}
\begin{table}[htbp]

\centering
\caption{Effectiveness of RSM for prompt tuning on the {DTD}.}
\resizebox{0.34\textwidth}{!}{%
\begin{tabular}{cccc}
\toprule
Method & Base & New & HM \\
\midrule
MaPLe & 81.75 & 53.50 & 64.67 \\
MaPLe + RSM & 82.29 & 55.56 & 66.33 \\
\cmidrule(lr){2-4}
\(\Delta\) &  {+0.54} &  {+2.06} &  {+1.66} \\
\midrule
PromptSRC & 82.83 & 61.19 & 70.39 \\
PromptSRC + RSM & 82.48 & 63.41 & 71.70 \\
\cmidrule(lr){2-4}
\(\Delta\)  &  {-0.35} &  {+2.22} &  {+1.31} \\
\bottomrule
\end{tabular}
}
\label{PTT}
\end{table}

Having evaluated the generalization performance of the prompt tuning approaches augmented with our proposed RSM, Table~\ref{PTT} represents an analysis for adaptability on the prompt tuning paradigm under supervised training manner. In specific, our method significantly improves the performance of MaPLe, yielding gains of 0.54\%, 2.06\%, and 1.66\% on Base, New, and HM, respectively. 
For PromptSRC, improvements of 2.22\% and 1.31\% can be also observed on New and HM.
These findings suggest that the proposed RSM is broadly applicable across diverse vision-language knowledge distillation model frameworks and consistently contributes to improved generalization performance on a wide range of downstream tasks.

\subsubsection{Effect on learnable projection}
\begin{table}[htbp]
\centering
\caption{Effect of the learnable projection on performance. Here, B and N stand for Base and New class, respectively.}

\setlength{\tabcolsep}{3pt}
\resizebox{0.46\textwidth}{!}{%
\begin{tabular}{c|ccc|ccc|ccc}
\toprule
{Learn.} & \multicolumn{3}{c|}{DTD} & \multicolumn{3}{c|}{EuroSAT} & \multicolumn{3}{c}{Average} \\
Proj.& B & N & HM & B & N & HM & B & N & HM \\
\midrule
 & 83.76 & 67.43 & 74.71 & 74.79 & 60.68 & 67.00 & 82.84 & 76.75 & 79.68 \\
\checkmark & 84.15 & 69.16 & 75.92 & 79.94 & 65.03 & 71.72 & 83.31 & 77.30 & 80.19 \\
\cmidrule(lr){2-10}
$\Delta$ & +0.39 & +1.73 & +1.21 & +5.15 & +4.35 & +4.72 & +0.46 & +0.54 & +0.51 \\
\bottomrule
\end{tabular}
}
\label{learnable}
\end{table}
Table~\ref{learnable} evaluates the effectiveness of the learnable projection used in the reconfigured latent space across two representative datasets, along with the average performance over all 11 datasets.
For example, the average Base, New, and HM scores improve from 82.84\%, 76.75\% and 79.68\% to 83.31\%, 77.30\% and 80.19\%, respectively. 
Note that a substantial improvement is observed on EuroSAT, with significant gains of 5.15\%, 4.35\% and 4.72\% for Base, New, and HM, respectively.

\begin{figure}[htp]

\centering
  \includegraphics[width=0.95\linewidth]{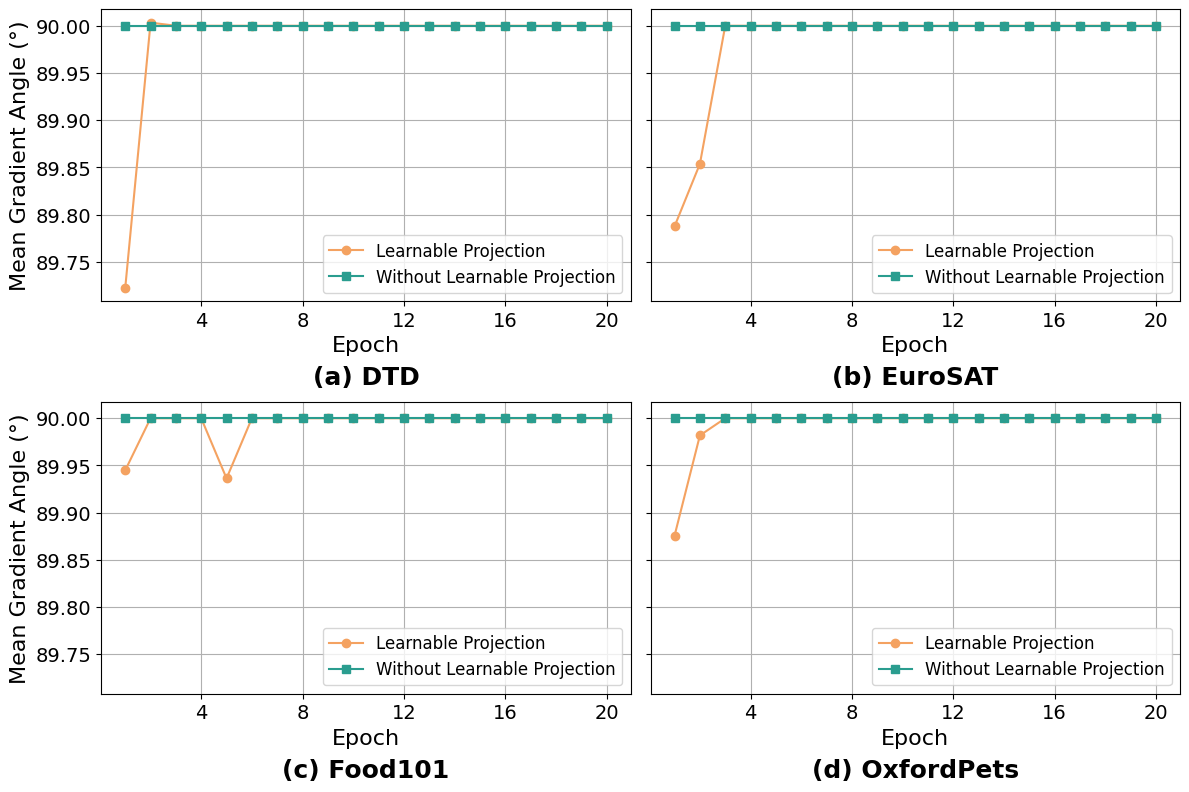}
    \captionsetup{labelsep=period }
\captionsetup{justification=raggedright}
  \caption{Gradient angle between the KL loss and the infomation entropy minimization loss.}
  \label{aaaagradient}
\end{figure}
Notably, Figure~\ref{aaaagradient} demonstrate that information entropy minimization begins to interact with the training dynamics when a learnable prompt is introduced, as evidenced by the gradient angle between the KL loss and the entropy loss dropping below 90 degrees across four datasets, including DTD, EuroSAT, Food101, and OxfordPets.
In contrast, without the learnable projection, the gradient angle stays around 90 degrees, indicating orthogonality and weak interaction between the two training objectives.

This finding aligns with our motivation to project multi-modal features into a shared manifold, thereby enabling effective feature compression and enhancing intra-class determinacy.
\begin{figure}[hbp]
\centering
\includegraphics[width=0.98\linewidth]{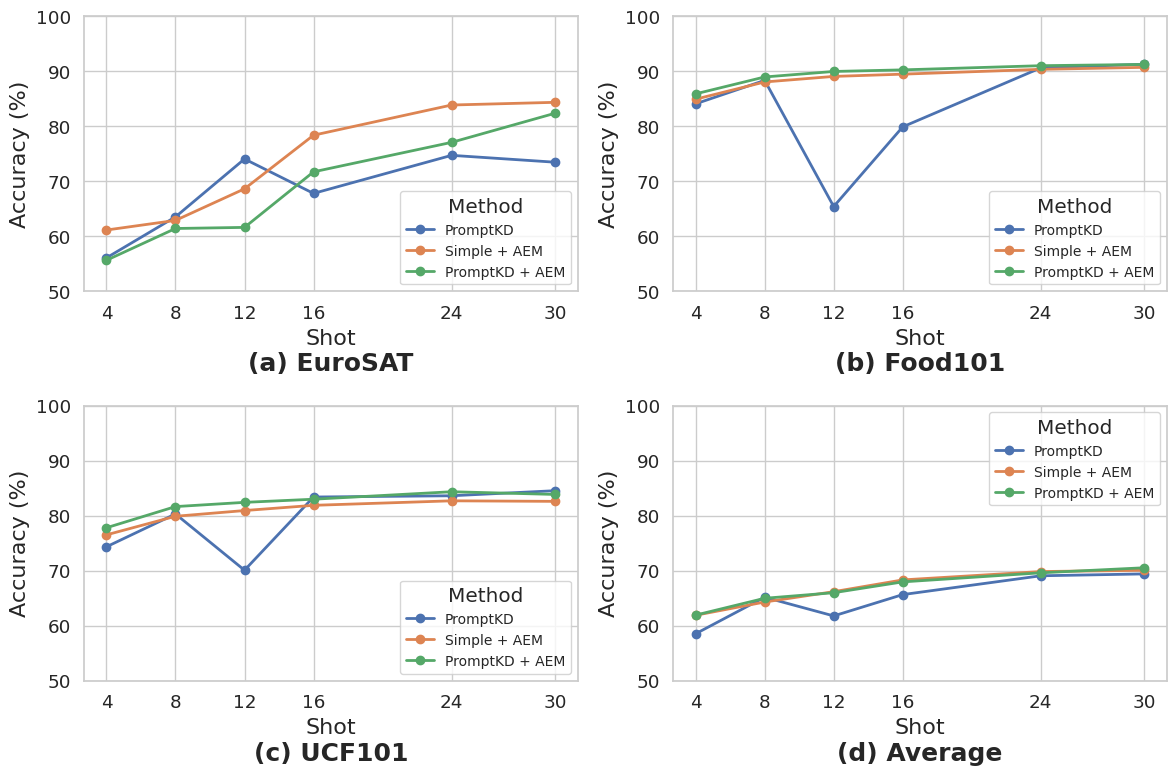}
  \captionsetup{labelsep=period }
\captionsetup{justification=raggedright}
\caption{Performance comparison across three datasets and their averages under different shot settings.}
\label{across_shot}
\end{figure}

\begin{figure*}[htbp]
  \centering
  \includegraphics[width=0.93\linewidth]{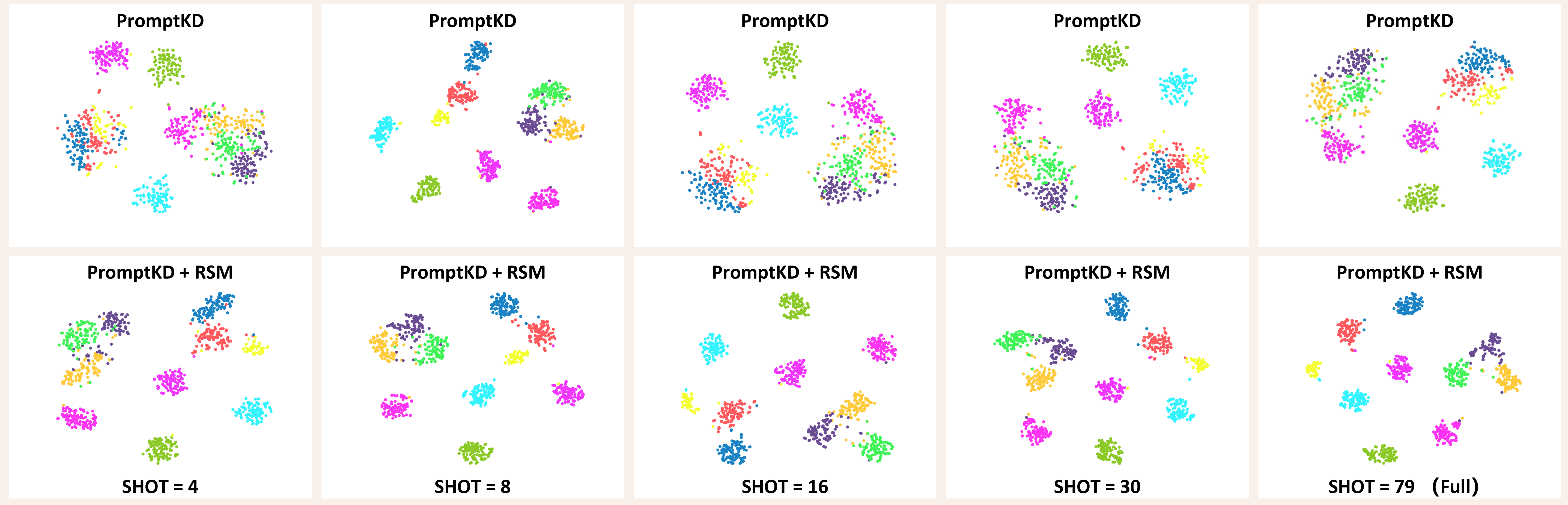}
    \captionsetup{labelsep=period }
\captionsetup{justification=raggedright}
  \caption{t-SNE visualization on the OxfordPets dataset across different shot settings.}
  \label{sne}
\end{figure*}


\subsubsection{Effect of Different Shot Settings Across Methods}
Figure~\ref{across_shot} illustrates the performance curves across different shot settings for three methods, PromptKD, Simple + RSM and PromptKD + RSM.
Notably, the methods incorporating our proposed RSM exhibit consistent improvements as the number of training shots increases. 
Such results clearly suggest that RSM encourages the model to effectively learn from the teacher, even in the limited data conditions.

\subsection{Visualization}
To facilitate understanding of the superior generalization improvement achieved by the proposed RSM, t-SNE visualization is conducted for both PromptKD and PromptKD + RSM under different training shot settings. 
As shown in Fig.~\ref{sne}, PromptKD + RSM exhibits more distinct discriminative clustering compared to the original PromptKD on the OxfordPets dataset.
Notably, clear class boundaries are consistently observed in PromptKD + RSM across all training shots, whereas disordered representations appear in PromptKD at 4-shot, 16-shot, 30-shot, and full-shot settings.


\section{Conclusion}
Knowledge distillation has re-emerged as a common strategy for transferring knowledge from large-scale vision-language models to smaller ones across diverse downstream tasks.
However, this paradigm exhibits limited generalization to samples with ambiguous or boundary-adjacent representations, often requiring large-scale training data to accurately delineate semantic boundaries in the representation space.
Such limitations constrain the broader adoption of knowledge distillation in real-world scenarios.
In this study, we propose \textbf{A}ligned \textbf{M}anifold \textbf{E}ntropy for Robust Vision-Language Distillation (AME), a method that enforces intra-class compression of cross-modal representations by applying entropy minimization over a reconfigured shared manifold.
This mechanism enables the model to structurally compress cross-modal representations during training, including those that are ambiguous or lie near decision boundaries, thereby enhancing training robustness and preserving generalization under data scarcity.
Extensive experiments and theoretical analysis clearly indicate that AME facilitates the distillation process with both improved robustness and superior generalization performance.
Crucially, this work proposes a lightweight, plug-and-play method that paves the way for broader application of vision-language knowledge distillation in real-world tasks.

\appendix
\section{Proof}
\begin{proof}[Proof of Corollary~\ref{co}]
Given the definition of conditional mutual information:
\begin{align}
I(W; V \mid S) = H(W \mid S) + H(V \mid S) - H(W, V \mid S),
\end{align}
where $H(W \mid S)$ and $H(V \mid S)$ denote the conditional entropies of the text and image embeddings, respectively.

The KL divergence term $\mathrm{KL}(P \| Q)$ minimizes the discrepancy between distributions $P$ and $Q$, thereby increasing $I(W; V \mid S)$ by aligning the probabilistic outputs of the teacher ($T$) and student ($V$).
In addition, the entropy term $H(M)$ serves as an explicit regularizer on the shared manifold. Since $H(M)$ approximates $H(W, V \mid S)$, minimizing $H(M)$ effectively reduces $H(W, V \mid S)$  and thus increases the conditional mutual information, strengthening the dependency between text and image embeddings, as follows:
\begin{align}
I(W; V \mid S) = H(W \mid S) + H(V \mid S) - H(W, V \mid S)\downarrow.
\end{align}
Here, since the total loss mainly targets the joint alignment between $T$ and $V$, rather than modifying the marginal distributions, we assume that $H(W \mid S)$ and $H(V \mid S)$ remain approximately constant.
\end{proof}


\begin{proof}[Proof of Corollary~\ref{mutual InB}]
Given the setting of theorem, assume:
\begin{itemize}
    \item The distillation KL loss $\mathcal{L}_{\mathrm{KD}}$ increases the dependency between model parameters $\theta$ and the training data $\mathcal{S}$, such that the mutual information is defined as $I(\theta; \mathcal{S})$.    
    \item The entropy regularization term $\omega H(M)$ constrains the expressiveness of the reconfigured latent space $M$.  
\end{itemize}

Since $M$ is a deterministic function of $(\theta, \mathcal{S})$, the data processing inequality gives:
\begin{align}
I(M; \mathcal{S}) \leq H(M).
\end{align}
By applying the chain rule of mutual information, we obtain:
\begin{align}
I(\theta; \mathcal{S}) = I(M; \mathcal{S}) + I(\theta; \mathcal{S} \mid M)\leq H(M) + \epsilon\leq \delta + \epsilon,
\end{align}
where $\epsilon = I(\theta; \mathcal{S}) - I(M; \mathcal{S})$ quantifies the residual dependency not captured by $M$. Therefore, the generalization error bound becomes:
\begin{align}
\mathbb{E}_{\mathcal{S}, \theta} \bigl[ L(\theta; \mathcal{D}) - L(\theta; \mathcal{S}) \bigr]
\;\leq\;
\sqrt{\dfrac{2 (\delta + \epsilon)}{n}}.
\end{align}
\end{proof}

\newpage 
\section*{Disclosure of GenAI}
Generative AI tools are utilized solely for polishing advice and grammar correction during the preparation of this paper. All content, such as research motivation, questions, experiments, and conclusions, is original contributions of the authors.

\end{document}